\documentclass[11pt]{article}

\usepackage{acl}

\usepackage{graphicx}
\usepackage{etoolbox}
\usepackage{spverbatim}
\usepackage{booktabs} 

\AtBeginEnvironment{spverbatim}{\small}

\usepackage{times}
\usepackage{latexsym}
\usepackage{amsmath}
\usepackage{makecell}

\usepackage[T1]{fontenc}

\usepackage{amssymb} 
\usepackage{pifont} 
\newcommand{\cmark}{\ding{51}} 
\newcommand{\xmark}{\ding{55}} 
 \usepackage{ulem}
\usepackage[utf8]{inputenc}
\usepackage{microtype}

\title{\textit{CharacterEval}: A Chinese Benchmark for Role-Playing \\ Conversational Agent Evaluation}

\author{
    Quan Tu\textsuperscript{$\spadesuit$}, \ 
    Shilong Fan\textsuperscript{$\clubsuit$}, \ 
    Zihang Tian\textsuperscript{$\spadesuit$}, \ 
    Rui Yan\textsuperscript{$\spadesuit$}\thanks{Corresponding author: Rui Yan (ruiyan@ruc.edu.cn).} \\
$^\spadesuit$Gaoling School of Artificial Intelligence, Renmin University of China \\
$^\clubsuit$School of Artificial Intelligence, Beijing University of Posts and Telecommunications \\
{\tt $^{\spadesuit}$\{quantu,tzh2003,ruiyan\}@ruc.edu.cn}\\
{\tt $^{\clubsuit}$fansl@bupt.edu.cn}\\
}

\begin{document}
\maketitle
\begin{abstract}
Recently, the advent of large language models (LLMs) has revolutionized generative agents. 
Among them, Role-Playing Conversational Agents (RPCAs) attract considerable attention due to their ability to emotionally engage users.
However, the absence of a comprehensive benchmark impedes progress in this field. 
To bridge this gap, we introduce \textit{CharacterEval}, a Chinese benchmark for comprehensive RPCA assessment, complemented by a tailored high-quality dataset.
The dataset comprises 1,785 multi-turn role-playing dialogues, encompassing 11,376 examples and featuring 77 characters derived from Chinese novels and scripts. 
It was carefully constructed, beginning with initial dialogue extraction via GPT-4, followed by rigorous human-led quality control, and enhanced with in-depth character profiles sourced from Baidu Baike.
\textit{CharacterEval} employs a multifaceted evaluation approach, encompassing thirteen targeted metrics on four dimensions. 
To facilitate the convenient evaluation for these subjective metrics in \textit{CharacterEval}, we further developed CharacterRM, a role-playing reward model based on human annotations, which has a higher correlation with human judgment compared to GPT-4.
Comprehensive experiments on \textit{CharacterEval} demonstrate that Chinese LLMs exhibit more promising capabilities than GPT-4 in Chinese role-playing conversation. 
Source code, data source and reward model will be publicly accessible at~\url{https://github.com/morecry/CharacterEval}.

\end{abstract}

\section{Introduction}




The development of large language models (LLMs) has marked the beginning of a new era in conversational AI~\cite{zhao2023survey, chang2023survey}, and opened up a wide range of application possibilities, particularly in agent-based interactions~\cite{park2023generative,wang2023survey,gao2023s}. 
The automated agents, equipped with the emerging capabilities of LLMs such as planning~\cite{silver2022pddl,ge2023openagi,song2023llm}, reasoning~\cite{wei2022chain,wang2022self}, and in-context learning~\cite{dong2022survey,brown2020language}, can perform complex tasks for humans without any supervision.  
Among the diverse agents, the Role-Playing Conversational Agent (RPCA), designed to offer emotional value instead of the productivity, attracts amount of interest. 
\begin{figure*}[!htbp]
    \centering
    \includegraphics[width=0.99\linewidth]{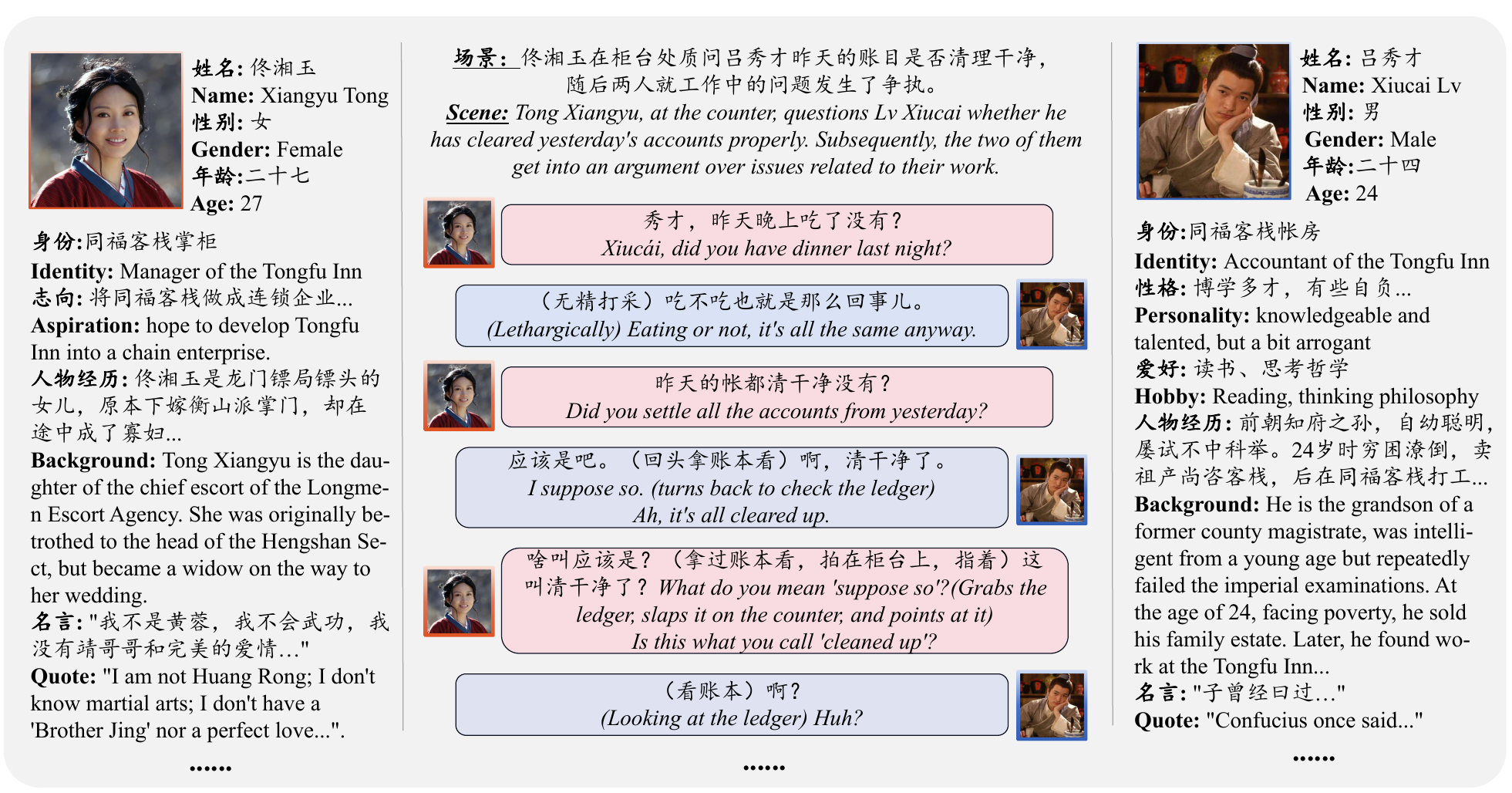}
    \caption{An example of the \textit{CharacterEval}, including the dialogue, scene and character's profile.}
    \label{fig:example}
\end{figure*}

RPCA represents a unique category within the realm of conversational agents, distinguished by their capability for immersive interaction~\cite{li2023chatharuhi}.
Different from traditional dialogue systems, which typically focus on chit-chat~\cite{yan2022deep}, knowledge-based~\cite{chen2020bridging}, personalized~\cite{zheng2019personalized} and empathetic dialogue~\cite{ma2020survey}, RPCAs engage users in dynamic scenarios, where LLM agents are assumed as specific characters or roles, often derived from existing composition such as novels, films, cartoons, and games. 
The development of connections between fictional characters and humans has the potential to not only deepen the impact of cultural works but also improve human engagement. 
Furthermore, RPCAs hold significant application value in their ability to offer emotional value to users, positioning fictional characters as virtual friends. 
The multifaceted nature of RPCAs has sparked considerable attention, leading to a surge in both research~\cite{shao2023character, wang2023rolellm, tu2023characterchat, zhou2023characterglm} and application development (e.g., Character AI\footnote{https://beta.character.ai}, Tongyi Xingchen\footnote{https://xingchen.aliyun.com/xingchen} and Glow\footnote{https://www.glowapp.tech/}). 
However, these implementations of RPCAs vary significantly in both approach and objectives, presenting a challenge in systematically assessing and comparing their capabilities. Therefore, we propose the \textit{CharacterEval}, a Chinese role-playing conversation benchmark for advancing RPCA development.

To develop a benchmark, the primary problem is the construction of a dataset. While there are existing datasets~\cite{shao2023character, wang2023rolellm, tu2023characterchat, zhou2023characterglm, chen2023large}, their quality are concerning, which are either generated by LLMs or suffering from significant noise due to the extractive methods. These limitations render the evaluation results unreliable for the RPCA's actual capabilities.
To address it, we constructed a Chinese role-playing conversation dataset comprising 1,785 multi-turn role-playing dialogues, encompassing 11,376 examples and 77 leading characters, drawn from diverse Chinese novels and scripts. 
Our process began with the collection of well-known sources across various genres. After that, GPT-4 was employed to extract dialogue scenes, utterances, and behaviors of the leading roles of these sources. 
Following basic preprocessing and the removal of dialogues with fewer turns, we invited annotators to assess the quality of the dialogues. Their task was to identify and retain high-quality dialogues, while discarding those of lower quality. Additionally, we crawled detailed character profiles from Baidu Baike\footnote{https://baike.baidu.com/}, composing a comprehensive dataset for RPCA evaluation. The example from the dataset is as Figure~\ref{fig:example} shows.

Otherwise, role-playing conversation is a complicated task that requires not only mimicking a character's behavior and utterance but also maintaining the character's knowledge, as well as the excellent multi-turn ability.
Considering this, we proposed a multifaceted evaluation approach including thirteen specific metrics on four dimensions for a fair and thorough assessment of RPCAs, 
Our evaluation approach considered the conversational ability, character consistency, role-playing attractiveness, and utilized a personality back-testing method to evaluate the personality accuracy of a RPCA.
To assess conversational ability, we measured the conversational fluency, coherence, and consistency at both the sentence and conversation levels~\cite{chen2017survey}.
Character consistency is the most crucial in role-playing conversation. 
Hence, we evaluated knowledge and persona consistency to measure how vividly an RPCA can simulate a character. This involves assessing knowledge exposure, accuracy, and hallucination for knowledge consistency, and evaluating behavior and utterance consistency for persona consistency.
Considering that RPCAs are entertainment-oriented, role-playing attractiveness is also a important elements. We assessed this through human-likeness, communication skill, expression diversity, and empathy.
Finally, we introduced personality back-testing. With the collected Myers-Briggs Type Indicator(MBTI)~\cite{myers1962myers} personality types as a reference, we let RPCAs do the MBTI assessment and calculate the MBTI accuracy (personality back-test) as implemented in~\citet{wang2023does}. 

For convenient re-implementation, we invited 12 annotators to score responses generated by different models for the subjective metrics in our evaluation system. Based the human judgments, we developed a role-playing reward model—CharacterRM, whose correlation with human could surpass state-of-the-art LLM GPT-4.
On \textit{CharacterEval}, We conducted comprehensive evaluations for existing LLMs, encompassing both open- and closed-source models. 
Experimental results shows the broad prospect of existing Chinese LLM while GPT-series models do not take the predominance in Chinese role-playing conversation. 

In summary, our contributions of are as follows:
\begin{itemize}
    \item We create a large-scale, high-quality dataset for RPCA evaluation, consisting of 1,785 multi-turn role-playing dialogues, 11,376 example, featuring 77 leading characters from diverse Chinese novels and scripts. 
    \item We propose \textit{CharacterEval}, a new benchmark for RPCAs, which contain comprehensive set of evaluation principles, encompassing thirteen specific metrics on four dimensions. 
    \item We develop CharacterRM, a role-playing reward model for evaluating RPCAs in several subjective metrics, achieving the better performance than GPT-4 on correlation with human.  
    \item We conducted thorough evaluations of existing LLMs on \textit{CharacterEval}, including open- and closed-source, and derived valuable findings from the results.
\end{itemize}

\section{Related Work}

\subsection{Knowledge-based Dialogue}
Knowledge-based dialogue systems integrate external knowledge resources, such as knowledge graphs or unstructured documents, into dialogue systems~\cite{zhao2020low, li2020zero}. Recent efforts have focused on improving the understanding and utilization of knowledge within these dialogues. For instance, \citet{xue2023improving} introduced K-DIAL, which incorporates additional Feed-Forward Network (FFN) blocks into Transformers~\cite{vaswani2017attention} to enhance factual knowledge expression and consistency in dialogue. Similarly, \citet{chen2020bridging} proposed a knowledge distillation-based training strategy to optimize the knowledge selection decoder. While these methods significantly advance knowledge selection and utilization, they primarily address general knowledge. Role-playing dialogues, however, demand a more intricate approach, encompassing personalized knowledge, style, behavior, etc.

\subsection{Personalized Dialogue}
Personalized dialogue systems, which generate responses based on specific personas, represent another relevant area of research~\cite{den2019reinforcement,zhong2022less}. \citet{zheng2019personalized} pioneered this field by creating the first large-scale personalized dialogue dataset, complete with persona labels. This dataset has spurred further advancements in the field. Additionally, \citet{zheng2020pre} developed a pre-trained personalized dialogue model, which could generate coherent responses using persona-sparse dialogue. Although these studies begin to explore persona in dialogue, the personal profiles they utilize are typically limited to short-term, person-related information like name, age, and location, which are considered as personalized knowledge in essence.

\subsection{Character-based Dialogue}
The most closely related research to this work involves recent developments in character-based dialogue systems, which aim to mimic the behavior and utterance style of specific characters~\cite{shao2023character, wang2023rolellm, zhou2023characterglm}. \citet{shao2023character} gathered character profiles from Wikipedia and generated character-based dialogues by prompting ChatGPT~\cite{chatgpt}. \citet{wang2023rolellm} used GPT-4 to create character descriptions and developed detailed instructions for prompting ChatGPT to produce character-based dialogues. However, these approaches primarily rely on ChatGPT's generative capabilities and may not accurately reflect the true personality of the characters. \citet{li2023chatharuhi} address this by extracting role-playing dialogues from novels, scripts, and games, which better preserve the characters' original traits. Despite this, their approach suffers from a lack of human-in-the-loop refinement and a scarcity of multi-turn dialogues in the dataset. Otherwise,~\citet{chen2023large} develop a role-playing dataset focused on \textit{Harry Potter}. However, the scarcity of diversity makes it hard to comprehensively evaluate the generalized RPCA.

\section{Problem Formulation}
\label{sec:formulation}
The Role-Playing Conversational Agent (RPCA) is designed to engage in conversations with users by emulating specific characters. These characters are defined by their knowledge, behavior, and style of the response. To achieve this, the RPCA utilizes a character profile, denoted as $P$, and the current dialogue context, represented as $C_n = [q_1, r_1, q_2, r_2, \dots, q_n]$. Here, $q_i$ and $r_i$ correspond to the $i$-th question and response in the dialogue, respectively. The goal for the RPCA is to generate a response $r_n$ that is consistent with the character's profile, which can be represented as:
\begin{equation}
r_n = \texttt{RPCA}(C_n, P),
\end{equation}
where $r_n$ is composed of two elements: behavior and utterance. The behavior aspect is enclosed in brackets and provides detailed description about the character's actions, expressions, and tone. 
This separation allows for a fine-grained evaluation of the RPCA's ability to not only generate appropriate utterance but also the unique behavioral traits.

\section{Data Collection}
In this section, we detail the methodology for constructing the character-based, multi-turn dialogue dataset with high quality. Prior to initiating data collection, adherence to the following four principles is important:

\begin{itemize}
    \item \textbf{Fidelity to Source Material:} It is crucial that all dialogues are in line with the original works, ensuring the character authenticity.
    \item \textbf{Diversity in Distribution:} The dataset must encompass a wide range of scenarios to thoroughly assess the role-playing capabilities. 
    \item \textbf{Multi-Turn Feature:} The dataset should predominantly consist of multi-turn dialogues, rather than being limited to single-turn ones.
    \item \textbf{Human-in-the-Loop:} Active human involvement is necessary to guarantee the quality, as reliance solely on LLMs is insufficient.
\end{itemize}

The pipeline of data collection includes four steps: plot division, dialogue extraction, quality filtering, and human annotation.

\textbf{Plot Division:} The plots in narrative text such as novels and scripts are extremely complex, making it challenging to divide the text into meaningful chunks. Using the sentence tokenization tool, without considering semantics, will result in breaking a conversation mid-way. To address this, we first employ GPT-4 to identify plot twists—sentences that signify the end of a continuous plot. These plot twists are then used to segment the text into chunks, each containing a complete plot.

\textbf{Dialogue Extraction:} Once we have the plot chunks, GPT-4 is utilized again, this time to extract role-playing dialogues. We design prompts for GPT-4 to perform information extraction, preserving characters' utterances, behaviors, and scenes from the plots.

\textbf{Quality Filtering:} Dialogues in novels and scripts often involve more than two characters. Simply retaining dialogues between two characters and omitting others will distort the dialogue structure. Therefore, we opt to preserve dialogues following an \texttt{ABAB} pattern (dialogue between two characters) until a third character joins. This approach, while straightforward, helps maintain the original dialogue structure more effectively. Besides, we only keep the dialogue exceed five turns (six sentences) reserved, filtering the short dialogues.  

\textbf{Human Annotation:} Although LLMs have the capabilitie in performing basic information extraction tasks, the randomness still affect data quality. To mitigate this, we invite human annotators to assess the coherence and quality of dialogues and eliminate any problematic instances.






\section{Evaluation Metric}

\begin{figure}
    \centering
    \includegraphics[width=0.99\linewidth]{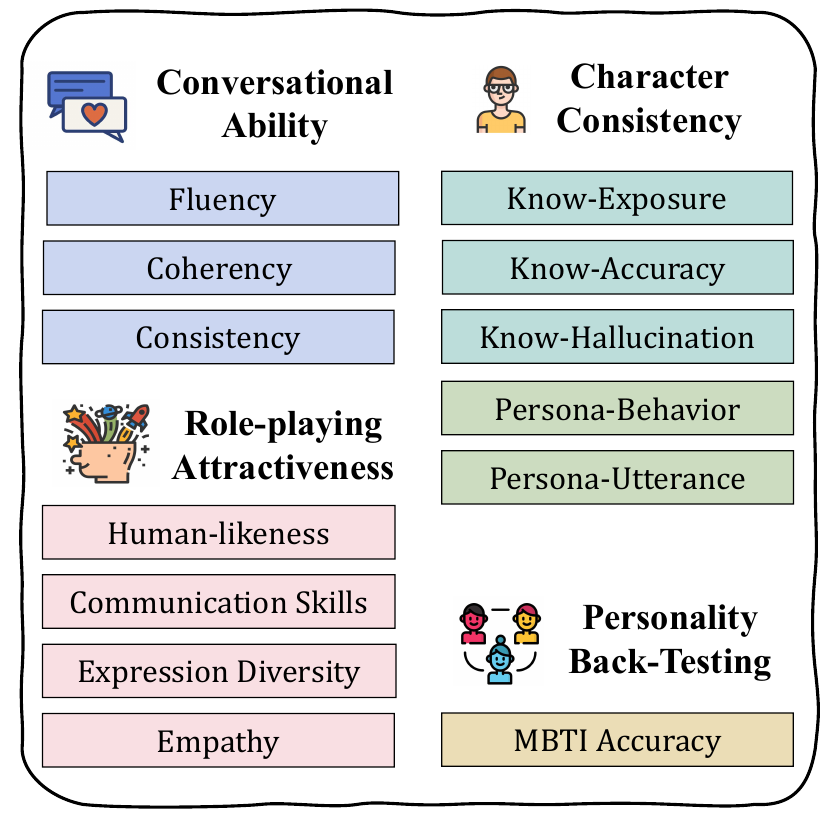}
    \caption{Evaluation system of \textit{CharacterEval}. ``Know-'' is the abbreviation of ``Knowledge''.}
    \label{fig:eval-sys}
\end{figure}
Different from traditional chatbots, we contend that RPCAs require a more comprehensive evaluation framework to assess their role-playing capabilities. As shown in Figure~\ref{fig:eval-sys}, we have devised a four-dimensional evaluation system, which are conversational ability, character consistency, role-playing attractiveness, and personality back-testing, including thirteen metrics.

\subsection{Conversational Ability}

The basic conversational ability is the first consideration in role-playing conversation. Inspired by previous neural metrics, which evaluate the responses based on well-trained neural models, we introduce the similar approach to assess the fundamental conversational abilities of RPCAs. We focus on three key objectives for generated responses: fluency, coherency, and consistency~\cite{zhang2021dynaeval, mesgar2020improving}.

\begin{itemize}
    \item \textbf{Fluency (Flu.)} measures the grammatical correctness of a response, indicating whether a response is readable and free from obvious grammatical errors.
    
    \item \textbf{Coherency (Coh.)} evaluates the topic relevance between the response and the context. Generally, when the user submits a query on a specific topic, an RPCA should respond following the topic instead of providing an irrelevant response.
    
    \item \textbf{Consistency (Cons.)} assesses the stability of RPCAs during a conversation. Responses of an RPCA should not contradict their own responses in previous turns.
\end{itemize}

\subsection{Character Consistency}
Character consistency plays an crucial role for evaluating the role-playing ability of the RPCAs. It will bring the most intuitive experience of users when the character consistency of RPCAs varies. Specifically, we evaluate the character consistency from two levels, knowledge consistency and persona consistency. The former evaluate if a RPCA could respond based on the character's knowledge, which include knowledge exposure, accuracy and hallucination metrics. The latter assess if a RPCA's reflection is in line with the character's personality, including the behavior and utterance metrics.

\begin{itemize}
    \item \textbf{Knowledge-Exposure (KE).} For assessing the informativeness of a response, it's crucial for an RPCA to reflect knowledge in its responses, as this support the subsequent evaluation of its knowledge expression capabilities.
    \item \textbf{Knowledge-Accuracy (KA).} Once the RPCA demonstrates the ability to generate responses with specific knowledge, it's important to assess whether this knowledge aligns with the character. The goal is for the RPCA to accurately generate response based on the knowledge from character's profile.
    \item \textbf{Knowledge-Hallucination (KH).} Drawing inspiration from recent studies on hallucinations in LLMs~\cite{rawte2023survey,zhang2023siren}, we include knowledge hallucination in the evaluation of role-playing dialogue. To enhance the user experience, the RPCA should maintain consistency with the character's identity and avoid responding to queries involving unknown knowledge.
    \item \textbf{Persona-Behavior (PB).} A character's behaviors, typically described within brackets, improve the embodied feeling of users by portraying the fine-grained actions, expressions, and tones. Consistent behavior is indicative for an effective RPCA.
    \item \textbf{Persona-Utterance (PU).} Alongside behavior, a character's speaking style is also important. Each character has unique expression habits. Therefore, the RPCA's utterances should align with these habits to adeptly mimic the character.
\end{itemize}



\subsection{Role-playing Attractiveness}

As a conversational agent in the entertainment field, it is essential for an RPCA to be sensitive to the user's emotions. Therefore, we introduce the character attractiveness dimension to assess the attraction of an RPCA during conversation. From the user's perspective, we consider four key dimensions: human-likeness, communication skills, expression diversity, and empathy.

\begin{itemize}
    \item \textbf{Human-Likeness (HL).} In the era of publicly available LLMs, these models often suffer from a perceived 'machine-like' quality in their responses. Most of LLMs, designed primarily for information seeking, tend to provide robotic and emotionless answers. However, in role-playing conversations, it is crucial for the RPCA to exhibit a more human-like persona to minimize user resistance.
    
    \item \textbf{Communication Skills (CS).} In human society, the ability to skilfully communicate, often referred to as Emotional Quotient (EQ), significantly influences an individual's likability. Accordingly, users are more likely to engage with an RPCA that demonstrates higher EQ, mirroring the popularity of individuals with strong communication skills in daily life.
    
    \item \textbf{Expression Diversity (ED).} The dialogues within \textit{CharacterEval} are sourced from existing novels, scripts, and various literary works, featuring characters with rich and diverse expressive abilities in both their behaviors and utterances. Therefore, an RPCA should strive to express this diversity in conversation to provide users with a more immersive experience.
    
    \item \textbf{Empathy (Emp.).} While the primary role of an RPCA is not that of an emotional counselor, its ability to express empathy can significantly impact its favorability of users. Evaluating empathy in role-playing conversations advance the RPCA to come across as a more warm and friendly conversational partner.
\end{itemize}


\subsection{Personality Back-Testing}
Following as the recent works on LLM personality testing~\cite{pan2023llms,huang2023chatgpt}, we conducted personality back-testing to assess the role-playing capability of the RPCA within the context of personality dimensions. In this study, we employed the Myers-Briggs Type Indicator (MBTI)~\cite{myers1962myers}, a well-established personality classification method.
To obtain the necessary labels, we collected MBTIs of characters featured in \textit{CharacterEval} from an archive website\footnote{https://www.personality-database.com/}, which hosts a substantial character's MBTIs. Using these MBTIs as ground truth, we evaluated the accuracy of the MBTI assessment~\footnote{https://www.16personalities.com/} of RPCAs.

\section{Experiment}
\subsection{Dataset Statistic}
\begin{table}[!htbp]
\small
  \centering
    \begin{tabular}{|l|c|c|}
     \hline
    \multicolumn{1}{|c|}{} & \multicolumn{1}{c|}{\textbf{Training}} & \multicolumn{1}{c|}{\textbf{Test}} \\
    \hline
    \# Characters & \multicolumn{2}{c|}{77} \\
    \hline
    \# Conversations & \multicolumn{2}{c|}{1,785} \\
    \hline
    Avg. Turns / Conv. & \multicolumn{2}{c|}{9.28} \\
    \hline
    Avg. Tokens / Conv. & \multicolumn{2}{c|}{369.69} \\
    \hline
    \# Examples & 6,811 & 4,564 \\
     \hline
    \end{tabular}%
    \caption{The statistic of \textit{CharacterEval} dataset.}
  \label{tab:statistic}%
\end{table}%
We split our \textit{CharacterEval} to training set and test set based on examples instead of conversations, where an example is composed by a tuple $(Character, Context, Response).$ The statistic of the dataset is as Table~\ref{tab:statistic} shows. The statistic demonstrates the multi-turn property of \textit{CharacterEval}, satisfying the evaluation of RPCAs performance at longer turns.

\subsection{Experimental Setting}


\begin{table}[h]
\small
\centering
\begin{tabular}{lllll}
\toprule
\textbf{Metric}                & \textbf{Char-RM} & \textbf{1-shot} & \textbf{2-shot} & \textbf{3-shot} \\ \midrule
Flu.            &     \textbf{0.613}             & 0.475           & 0.571           & 0.560           \\
Coh.            &     \textbf{0.607}             & 0.493           & 0.577           & 0.604           \\
Cons.           &     \textbf{0.573}             & 0.563           & 0.484           & 0.483           \\ \hline
KE              &     \textbf{0.509}             & 0.241           & 0.332           & 0.407           \\
KA              &     \textbf{0.336}             & 0.239           & 0.182           & 0.187           \\
KH              &     \textbf{0.411}             & 0.377           & 0.380           & 0.332           \\
PB              &     \textbf{0.879}             & 0.253           & 0.305           & 0.244           \\
PU              &     0.472             & 0.394           & 0.432           & \textbf{0.563}           \\ \hline
HL              &     \textbf{0.497}             & 0.271           & 0.308           & 0.318           \\
CS              &     \textbf{0.686}             & 0.489           & 0.350           & 0.371           \\
ED              &     \textbf{0.765}             & 0.209           & 0.298           & 0.301           \\
Emp.            &     0.385             & \textbf{0.407}           & 0.403           & 0.371           \\ \hline
Overall         &     \textbf{0.631}             & 0.362           & 0.385           & 0.375           \\ \bottomrule
\end{tabular}
\caption{Pearson correlation coefficient~\cite{pearson1901liii} with human judgements of GPT-4 and our CharacterRM (abbr. Char-RM). We report the performance of GPT-4 under different setting: 1-shot, 2-shot and 3-shot. \textbf{Bold} indicates the highest score.}
 \label{table:coefficients}
\end{table}

\begin{table*}[!htbp]
\centering
\small
\begin{tabular}{|l|c|c|c|c|c|}
\hline
\textbf{Models} & \textbf{Specialized} & \textbf{Model Size} & \textbf{Open Source} &  \textbf{Primarily Language} & \textbf{Creator} \\
\hline
ChatGLM3 & \xmark & 6B & \cmark & zh & Tsinghua \& Zhipu AI \\
XVERSE & \xmark & 7B, 13B & \cmark & zh & XVERSE \\
Qwen  & \xmark & 7B, 14B & \cmark & zh & Alibaba Inc. \\
InternLM & \xmark & 7B, 20B & \cmark & zh & SenseTime \& Shanghai AI lab \\
Baichuan2 & \xmark & 7B, 13B & \cmark & zh & Baichuan Inc. \\
CharacterGLM & \cmark & undisclosed & \xmark  & zh & Tsinghua \& Lingxin \\
Xingchen & \cmark & undisclosed & \xmark & zh & Alibaba Inc. \\
MiniMax & \cmark & undisclosed & \xmark & zh & MiniMax Inc. \\
BC-NPC-Turbo & \cmark & undisclosed & \xmark & zh & Baichuan Inc. \\
GPT-3.5 & \xmark & undisclosed & \xmark& en & OpenAI \\
GPT-4 & \xmark & undisclosed & \xmark& en & OpenAI \\

\hline
\end{tabular}
\caption{LLMs evaluated in our experiments.}
\end{table*}

\textit{CharacterEval} employs a comprehensive set of fine-grained subjective metrics (twelve metrics in conversational ability, character consistency and role-playing attractiveness dimensions) to assess the multi-dimensional capabilities of an Role-Playing Conversation Agent (RPCA). However, it is important to note that a single evaluated example may not adequately represent all facets of RPCAs. Therefore, we introduce annotators to sparsely evaluate the performance matrix. This approach entails that each example in \textit{CharacterEval} is assessed using a subset of these subjective metrics, leading to more differentiated evaluation results. 

Then, based on these selected metrics for each example, we recruit 12 annotators to score responses generated by different models in a five-point scale. The human judgements are used to develop a role-playing reward model (CharacterRM), with Baichuan2-13B-base as the backbone. Experimental result shows that our CharacterRM exhibits a higher correlation with human than GPT-4, as Table~\ref{table:coefficients} shows. 
Although the performance of GPT-4 will improve with the number of demonstration increase, the cost of it makes evaluation hard to implement. Consequently, we utilize our CharacterRM for subsequent evaluation on subjective metrics. 
In personality back-test, we collect 54 ground MBTIs of characters in our dataset. The RPCAs should answer the MBTI questionnaires and then the accuracy  will be computed.

\subsection{Evaluated LLMs}

In this work, we assess the performance of 10 baselines with different parameters, encompassing both open-source and closed-source models. For the open-source models, we evaluate their \textbf{chat-version} instead of base-version.
For the closed-source models, we utilize their official APIs to conduct performance evaluations. 
Specifically, we employ the $\texttt{gpt-4}$ version as the GPT-4, and $\texttt{gpt-3.5-turbo-1106}$ version as GPT-3.5 in our experiments. 
Among the evaluated models, CharacterGLM, MiniMax, and Xingchen and BC-NPC-Turbo are tailored for role-playing conversations, while the remaining models are designed for general chat applications. 
Notably, GPT-4 and GPT-3.5 stands out as the only two models trained on the dataset primarily composed of corpus with English language. 
We consistently employ the same prompt for each model, with minor adjustments made only for closed-source models.



\subsection{Overall Performance}
\begin{figure}[!h]
    \centering
    \includegraphics[width=0.99\linewidth]{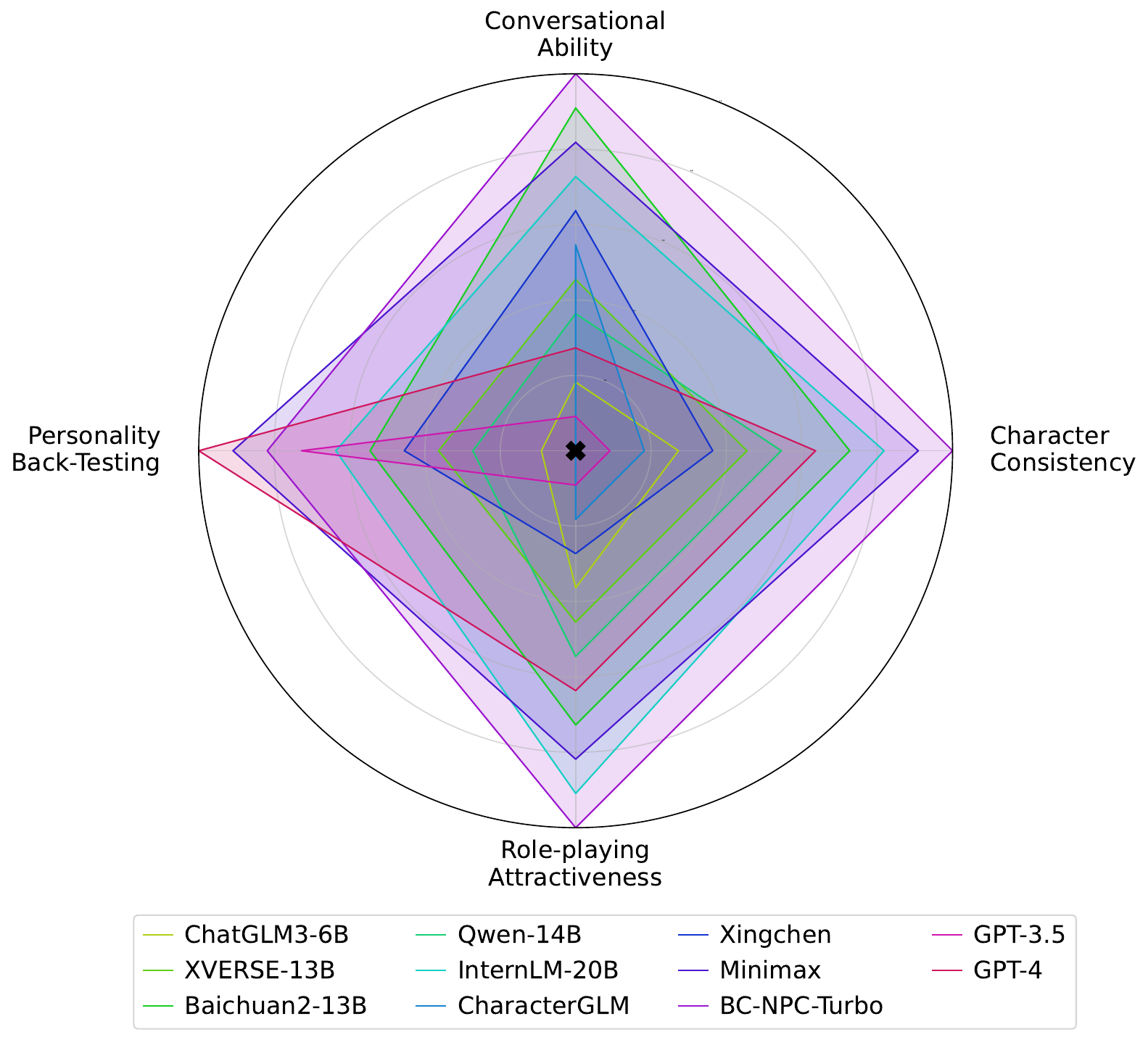}
    \caption{The comprehensive comparison of LLMs on four dimensions. Since CharacterGLM can not successfully complete personality back-testing, we mark the result using 'X' instead.}
    \label{fig:radar}
\end{figure}
The results across four dimensions are clearly illustrated in Figure~\ref{fig:radar}. BC-NPC-Turbo outperforms in three of these dimensions, whereas GPT-4 is distinguished in personality back-testing. Models specifically designed for role-playing dialogues, such as Xingchen, MiniMax, and BC-NPC-Turbo, demonstrate superior outcomes due to their targeted training. 
In the realm of open-source models, InternLM-20B and Baichuan2-13B show impressive potential. Despite lacking specialized customization for role-playing conversations, these models present commendable results in most evaluation dimensions. In contrast, GPT-4's effectiveness diminishes in Chinese role-playing conversations. Its primary training in English corpus limits the adaptability in complex role-playing scenarios and the deep understanding of Chinese culture.

\begin{table*}[!h]
\centering
\small


\begin{tabular}{|l|>{\centering\arraybackslash}m{0.91cm}>{\centering\arraybackslash}m{0.91cm}>{\centering\arraybackslash}m{0.91cm}>{\centering\arraybackslash}m{0.91cm}>{\centering\arraybackslash}m{0.91cm}>{\centering\arraybackslash}m{0.91cm}|>{\centering\arraybackslash}m{1.8cm}|}

\toprule

& \multicolumn{6}{c|}{\textbf{Character Consistency}}                                                                                                    & \textbf{Personality} \\ \cline{2-7}
& \multicolumn{1}{c}{KE} & \multicolumn{1}{c}{KA} & \multicolumn{1}{c}{KH} & \multicolumn{1}{c}{PB} & \multicolumn{1}{c}{PU} & \multicolumn{1}{c|}{Avg.} & \textbf{Back-Testing}        \\ \hline
ChatGLM3-6B        & \underline{2.016}            & 2.792                  & 2.704                  & 2.455                  & 2.812                  & 2.556                     & 0.532                \\
XVERSE-7B          & 1.834                  & 2.774                  & 2.763                  & 2.564                  & 2.887                  & 2.564                     & 0.62                 \\
Baichuan2-7B       & 1.813                  & 2.849                  & 2.929                  & \underline{2.830}            & 3.081                  & 2.700                     & 0.625                \\
Qwen-7B            & 1.956                  & 2.728                  & 2.633                  & 2.605                  & 2.780                  & 2.540                     & 0.606                \\
InternLM-7B        & 1.782                  & 2.800                  & 2.781                  & 2.719                  & 3.016                  & 2.620                     & 0.63                 \\
XVERSE-13B         & 1.977                  & 2.828                  & 2.862                  & 2.579                  & 2.915                  & 2.632                     & 0.63                 \\
Baichuan2-13B      & 1.802                  & 2.869                  & \underline{2.946}            & 2.808                  & 3.081                  & 2.701                     & 0.639                \\
Qwen-14B           & 1.988                  & 2.800                  & 2.811                  & 2.744                  & 2.900                  & 2.649                     & 0.62                 \\
InternLM-20B       & 1.945                  & \underline{2.916}            & 2.920                  & 2.753                  & 3.041                  & 2.715                     & 0.648                \\
CharacterGLM       & 1.640                  & 2.819                  & 2.738                  & 2.301                  & 2.969                  & 2.493                     & -                    \\
Xingchen           & 1.636                  & 2.768                  & 2.743                  & 2.772                  & 3.055                  & 2.595                     & 0.63                 \\
MiniMax            & 1.835                  & 2.910                  & 2.944                  & 2.774                  & \underline{3.125}            & \underline{2.718}               & \underline{0.685}          \\
BC-NPC-Turbo & 1.802                  & \textbf{2.964}         & \textbf{2.993}         & \textbf{2.910}         & \textbf{3.151}         & \textbf{2.764}            & 0.681                \\
GPT-3.5            & 1.716                  & 2.339                  & 2.212                  & 1.921                  & 2.316                  & 2.101                     & 0.653                \\
GPT-4              & \textbf{2.250}         & 2.855                  & 2.785                  & 2.721                  & 2.873                  & 2.697                     & \textbf{0.694}      
\end{tabular}
\begin{tabular}{|l|ccccccccc|}
\midrule
                   & \multicolumn{4}{c|}{\textbf{Conversational Ability}}                                                                          & \multicolumn{5}{c|}{\textbf{Role-playing Attractiveness}}                                                                                  \\ \cline{2-10} 
                   & Flu.                      & Coh.                      & Cons.                           & \multicolumn{1}{c|}{Avg.}           & HL                        & CS                        & ED                        & Emp.                      & Avg.                       \\ \hline
ChatGLM3-6B        & 3.269                     & 3.647                     & 3.283                           & \multicolumn{1}{c|}{3.399}          & 3.064                     & 2.932                     & 1.969                     & 2.993                     & 2.739                      \\
XVERSE-7B          & 3.393                     & 3.752                     & 3.518                           & \multicolumn{1}{c|}{3.554}          & 3.395                     & 2.743                     & 2.013                     & 2.936                     & 2.772                      \\
Baichuan2-7B       & \multicolumn{1}{l}{3.551} & \multicolumn{1}{l}{3.894} & \multicolumn{1}{l}{\underline{3.827}} & \multicolumn{1}{l|}{3.757}          & \multicolumn{1}{l}{3.670} & \multicolumn{1}{l}{2.728} & \multicolumn{1}{l}{2.115} & \multicolumn{1}{l}{2.984} & \multicolumn{1}{l|}{2.874} \\
Qwen-7B            & \multicolumn{1}{l}{3.187} & \multicolumn{1}{l}{3.564} & \multicolumn{1}{l}{3.229}       & \multicolumn{1}{l|}{3.327}          & \multicolumn{1}{l}{3.036} & \multicolumn{1}{l}{2.791} & \multicolumn{1}{l}{2.052} & \multicolumn{1}{l}{2.838} & \multicolumn{1}{l|}{2.679} \\
InternLM-7B        & 3.527                     & 3.823                     & 3.744                           & \multicolumn{1}{c|}{3.698}          & 3.546                     & 2.622                     & 2.070                     & 2.897                     & 2.784                      \\
XVERSE-13B         & 3.444                     & 3.811                     & 3.559                           & \multicolumn{1}{c|}{3.605}          & 3.319                     & \underline{2.939}               & 2.045                     & 3.018                     & 2.830                      \\
Baichuan2-13B      & \underline{3.596}               & 3.924                     & 3.864                           & \multicolumn{1}{c|}{\underline{3.795}}    & 3.700                     & 2.703                     & 2.136                     & \underline{3.021}               & 2.890                      \\
Qwen-14B           & 3.351                     & 3.765                     & 3.510                           & \multicolumn{1}{c|}{3.542}          & 3.354                     & 2.871                     & \underline{2.237}               & 2.970                     & 2.858                      \\
InternLM-20B       & 3.576                     & \textbf{3.943}            & 3.717                           & \multicolumn{1}{c|}{3.745}          & 3.582                     & 2.885                     & 2.132                     & \textbf{3.047}            & \underline{2.911}                \\
CharacterGLM       & 3.414                     & 3.717                     & 3.737                           & \multicolumn{1}{c|}{3.623}          & 3.738                     & 2.265                     & 1.966                     & 2.812                     & 2.695                      \\
Xingchen           & 3.378                     & 3.807                     & 3.754                           & \multicolumn{1}{c|}{3.646}          & 3.757                     & 2.272                     & 2.100                     & 2.799                     & 2.732                      \\
MiniMax            & \textbf{3.609}            & \underline{3.932}               & 3.811                           & \multicolumn{1}{c|}{3.784}          & \underline{3.768}               & 2.672                     & 2.150                     & 3.017                     & 2.902                      \\
BC-NPC-Turbo & 3.578                     & 3.898                     & \textbf{3.916}                  & \multicolumn{1}{c|}{\textbf{3.798}} & \textbf{3.836}            & 2.643                     & \textbf{2.336}            & 2.971                     & \textbf{2.946}             \\
GPT-3.5            & 2.629                     & 2.917                     & 2.700                           & \multicolumn{1}{c|}{2.749}          & 2.565                     & 2.422                     & 1.660                     & 2.526                     & 2.293                      \\
GPT-4              & 3.332                     & 3.669                     & 3.343                           & \multicolumn{1}{c|}{3.448}          & 3.143                     & \textbf{3.184}            & 2.153                     & 3.010                     & 2.873        \\\bottomrule             
\end{tabular}

\caption{Detailed evaluation results on \textit{CharacterEval}. The best performances are highlighted in \textbf{bold}, while sub-optimal ones are marked with \underline{underline}. It is notable that the score for CharacterGLM in personality back-testing is unavailable, hence it is replaced by a ``-''.}
\label{tab:detail_result}
\end{table*}
Significantly, GPT-3.5 demonstrates the weakest performance in \textit{CharacterEval}. Its tendency to generate overly safe responses, such as “\textit{I am just an AI assistant and cannot perform role-playing},” highlights its limitations for role-playing applications. 
This issue stems from the over alignment by RLHF~\cite{christiano2017deep}, making it unsuitable for dynamic role-playing interactions.


\subsection{Detailed Result}
The detailed performance across thirteen metrics is presented in Table~\ref{tab:detail_result}. 
Regarding conversational capabilities, BC-NPC-Turbo exhibits superior performance, evidenced by its excellent conversational consistency, as well as the comparative fluency and coherency. 
In contrasting open-source and closed-source models, it is difficult to declare a definitive winner in this dimension. However, when we compare the homogeneous models, such as Qwen-7B versus Qwen-14B, and XVERSE-7B versus XVERSE-13B, examining models of the same series, such as Qwen-7B versus Qwen-14B, and XVERSE-7B versus XVERSE-13B, it becomes obvious that an increase in the number of parameters can enhance conversational abilities.
In the category of models with fewer than 10 billion parameters, Baichuan2-7B and InternLM-7B demonstrate comparable competencies.
In the role-playing specialized models, MiniMax stands out for its performance and only fall behind  BC-NPC-Turbo. In constrast, GPT-4 and GPT-3.5 do not exhibit a marked superiority in this dimension.
Furthermore, it is posited that complex role-playing conversations and scenarios in Chinese might challenge the GPT series, potentially leading to their diminished performance.

In terms of character consistency, the most crucial aspect for role-playing conversations, BC-NPC-Turbo still leads significantly. It exhibits the highest accuracy in knowledge accuracy, minimal knowledge hallucinations, and consistent utterances and behaviors when acting as a character. 
Otherwise, MiniMax also shows notable performance, comparing with open-sourced model and remained closed-models. 
Once again, the GPT series, falls short compared to Chinese LLMs in this dimension. Nonetheless, it is important to acknowledge that GPT-4 excels in knowledge exposure, underlining its strengths in knowledge-intensive tasks. Despite this, in the realm of knowledge accuracy, particularly concerning the understanding of Chinese classical characters, GPT-4 does not exhibit distinct superiority.


Furthermore, BC-NPC-Turbo stands out in role-playing attractiveness, as demonstrated by its outstanding human-likeness and diverse expressions. 
As a state-of-the-art LLM, GPT-4 exhibits remarkable performance in communication skills, significantly surpassing other models. This reflects its powerful generalization ability, even in the Chinese role-playing scenario. 
Interestingly, InternLM-20B emerges as the leader in empathy, highlighting its unique potential to provide emotional support.

Similar conclusions are also observed in the personality back-test, where BC-NPC-Turbo, MiniMax, and GPT-4 demonstrate comparable levels of accuracy. 
In this particular dimension, the models are required to respond to multi-choice questions that are designed to reveal the underlying values of the roles they are portraying. 
Since this task does not demand extensive expression in the character's text style, GPT-4 exhibits the best performance. This result highlights their ability to accurately embody a character's fundamental personality traits and values.

\subsection{Robustness Analysis}

\begin{figure}[!htbp]
    \centering
    \includegraphics[width=0.99\linewidth]{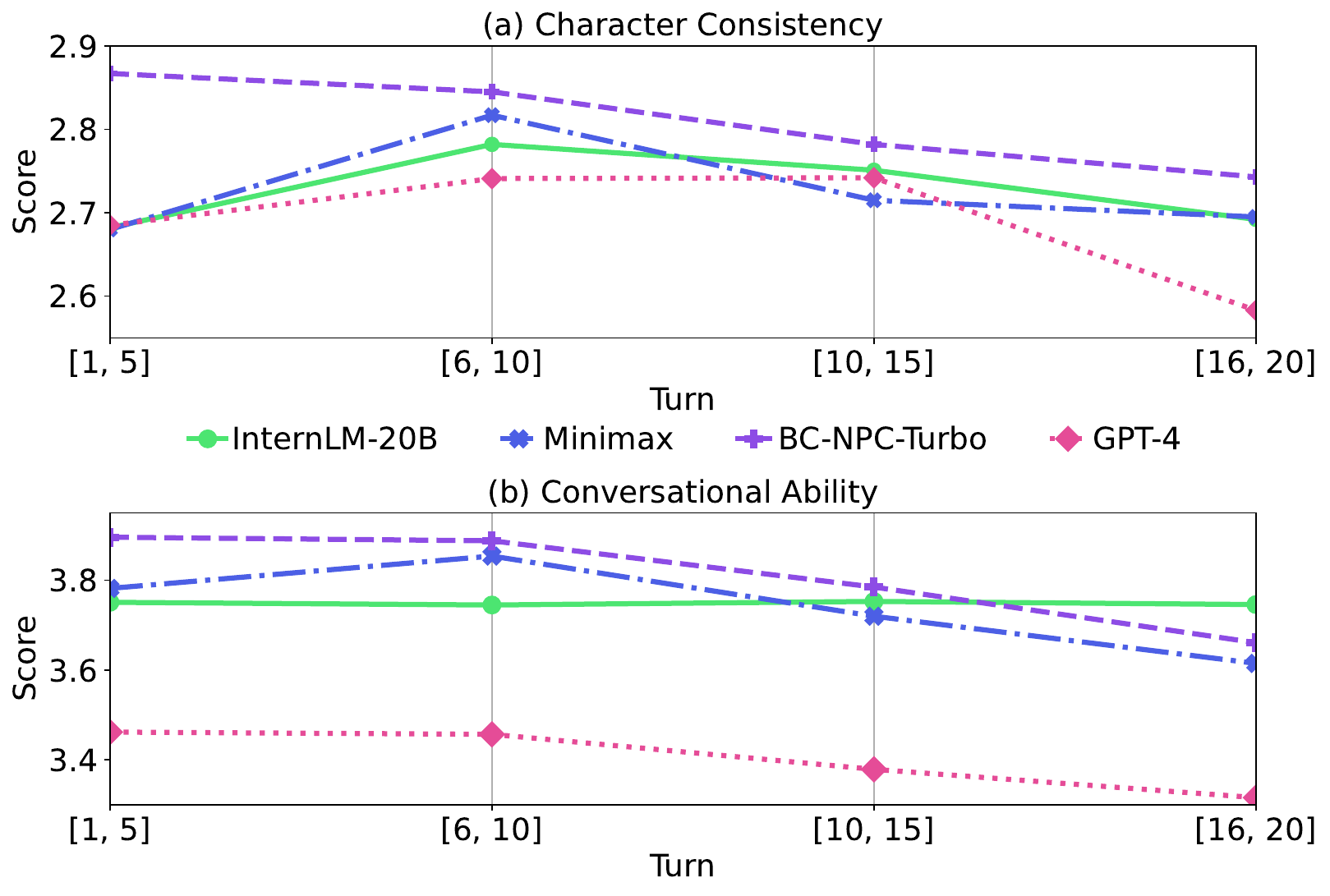}
    \caption{Model performance across the different stage of the conversation.}
    \label{fig:multiturn}
\end{figure}

To evaluate the robustness of RPCAs, we select a range of models—InternLM-20B, MiniMax, BC-NPC-Turbo, and GPT-4—for analysis. We aim to assess their effectiveness on different stages of a conversation. 
As illustrated in Figure~\ref{fig:multiturn}, there is a noticeable trend where most models demonstrate a decline in performance as conversations progress. Remarkably, InternLM-20B maintains consistent performance in terms of character consistency and conversational ability. 
This could be attributed to the fact that these models, primarily designed for role-playing, have not significantly focused on longer dialogue sequences. 
This oversight is likely due to the challenges associated with collecting extensive role-playing conversation data. 
Similarly, GPT-4 exhibits a declined trend under longer conversations, affected by the complex Chinese role-playing scenarios.
Our findings indicate that future advancements in RPCA development should focus on enhancing capabilities for longer conversational scenarios, ensuring more stable and consistent role-playing interactions.

\section{Conclusion}
In this work, we devote to build a comprehensive benchmark to evaluate recent Role-Playing conversational Agents (RPCAs). We introduce GPT-4 to extract the dialogues from the existing novels and scripts, proceeding with strict human filtering. 
After a series processing, we release a high-quality multi-turn role-playing dataset. 
Besides, we construct a comprehensive evaluating system to assess the multi-dimensional ability of RPCAs. 
We also collect human annotation to train a character-based reward model to measure the subjective metrics, for later convenient re-implementation.
Extensive experimental results indicate that Chinese LLMs entail more promising capabilities than GPT-4 in Chinese role-playing conversation.

\newpage
\bibliography{anthology,custom}

\begin{thebibliography}{42}
\expandafter\ifx\csname natexlab\endcsname\relax\def\natexlab#1{#1}\fi

\bibitem[{Brown et~al.(2020)Brown, Mann, Ryder, Subbiah, Kaplan, Dhariwal, Neelakantan, Shyam, Sastry, Askell et~al.}]{brown2020language}
Tom Brown, Benjamin Mann, Nick Ryder, Melanie Subbiah, Jared~D Kaplan, Prafulla Dhariwal, Arvind Neelakantan, Pranav Shyam, Girish Sastry, Amanda Askell, et~al. 2020.
\newblock Language models are few-shot learners.
\newblock \emph{Advances in neural information processing systems}, 33:1877--1901.

\bibitem[{Chang et~al.(2023)Chang, Wang, Wang, Wu, Zhu, Chen, Yang, Yi, Wang, Wang et~al.}]{chang2023survey}
Yupeng Chang, Xu~Wang, Jindong Wang, Yuan Wu, Kaijie Zhu, Hao Chen, Linyi Yang, Xiaoyuan Yi, Cunxiang Wang, Yidong Wang, et~al. 2023.
\newblock A survey on evaluation of large language models.
\newblock \emph{arXiv preprint arXiv:2307.03109}.

\bibitem[{Chen et~al.(2017)Chen, Liu, Yin, and Tang}]{chen2017survey}
Hongshen Chen, Xiaorui Liu, Dawei Yin, and Jiliang Tang. 2017.
\newblock A survey on dialogue systems: Recent advances and new frontiers.
\newblock \emph{Acm Sigkdd Explorations Newsletter}, 19(2):25--35.

\bibitem[{Chen et~al.(2023)Chen, Wang, Jiang, Cai, Li, Chen, Wang, and Li}]{chen2023large}
Nuo Chen, Yan Wang, Haiyun Jiang, Deng Cai, Yuhan Li, Ziyang Chen, Longyue Wang, and Jia Li. 2023.
\newblock Large language models meet harry potter: A dataset for aligning dialogue agents with characters.
\newblock In \emph{Findings of the Association for Computational Linguistics: EMNLP 2023}, pages 8506--8520.

\bibitem[{Chen et~al.(2020)Chen, Meng, Li, Chen, Xu, Xu, and Zhou}]{chen2020bridging}
Xiuyi Chen, Fandong Meng, Peng Li, Feilong Chen, Shuang Xu, Bo~Xu, and Jie Zhou. 2020.
\newblock Bridging the gap between prior and posterior knowledge selection for knowledge-grounded dialogue generation.
\newblock In \emph{Proceedings of the 2020 conference on empirical methods in natural language processing (EMNLP)}, pages 3426--3437.

\bibitem[{Christiano et~al.(2017)Christiano, Leike, Brown, Martic, Legg, and Amodei}]{christiano2017deep}
Paul~F Christiano, Jan Leike, Tom Brown, Miljan Martic, Shane Legg, and Dario Amodei. 2017.
\newblock Deep reinforcement learning from human preferences.
\newblock \emph{Advances in neural information processing systems}, 30.

\bibitem[{Den~Hengst et~al.(2019)Den~Hengst, Hoogendoorn, Van~Harmelen, and Bosman}]{den2019reinforcement}
Floris Den~Hengst, Mark Hoogendoorn, Frank Van~Harmelen, and Joost Bosman. 2019.
\newblock Reinforcement learning for personalized dialogue management.
\newblock In \emph{IEEE/WIC/ACM International Conference on Web Intelligence}, pages 59--67.

\bibitem[{Dong et~al.(2022)Dong, Li, Dai, Zheng, Wu, Chang, Sun, Xu, and Sui}]{dong2022survey}
Qingxiu Dong, Lei Li, Damai Dai, Ce~Zheng, Zhiyong Wu, Baobao Chang, Xu~Sun, Jingjing Xu, and Zhifang Sui. 2022.
\newblock A survey for in-context learning.
\newblock \emph{arXiv preprint arXiv:2301.00234}.

\bibitem[{Gao et~al.(2023)Gao, Lan, Lu, Mao, Piao, Wang, Jin, and Li}]{gao2023s}
Chen Gao, Xiaochong Lan, Zhihong Lu, Jinzhu Mao, Jinghua Piao, Huandong Wang, Depeng Jin, and Yong Li. 2023.
\newblock S3: Social-network simulation system with large language model-empowered agents.
\newblock \emph{arXiv preprint arXiv:2307.14984}.

\bibitem[{Ge et~al.(2023)Ge, Hua, Ji, Tan, Xu, and Zhang}]{ge2023openagi}
Yingqiang Ge, Wenyue Hua, Jianchao Ji, Juntao Tan, Shuyuan Xu, and Yongfeng Zhang. 2023.
\newblock Openagi: When llm meets domain experts.
\newblock \emph{arXiv preprint arXiv:2304.04370}.

\bibitem[{Huang et~al.(2023)Huang, Wang, Lam, Li, Jiao, and Lyu}]{huang2023chatgpt}
Jen-tse Huang, Wenxuan Wang, Man~Ho Lam, Eric~John Li, Wenxiang Jiao, and Michael~R Lyu. 2023.
\newblock Chatgpt an enfj, bard an istj: Empirical study on personalities of large language models.
\newblock \emph{arXiv preprint arXiv:2305.19926}.

\bibitem[{Li et~al.(2023)Li, Leng, Yan, Shen, Wang, MI, Fei, Feng, Yan, Wang et~al.}]{li2023chatharuhi}
Cheng Li, Ziang Leng, Chenxi Yan, Junyi Shen, Hao Wang, Weishi MI, Yaying Fei, Xiaoyang Feng, Song Yan, HaoSheng Wang, et~al. 2023.
\newblock Chatharuhi: Reviving anime character in reality via large language model.
\newblock \emph{arXiv preprint arXiv:2308.09597}.

\bibitem[{Li et~al.(2020)Li, Xu, Wu, Zhao, Zhao, and Tao}]{li2020zero}
Linxiao Li, Can Xu, Wei Wu, Yufan Zhao, Xueliang Zhao, and Chongyang Tao. 2020.
\newblock Zero-resource knowledge-grounded dialogue generation.
\newblock \emph{Advances in Neural Information Processing Systems}, 33:8475--8485.

\bibitem[{Ma et~al.(2020)Ma, Nguyen, Xing, and Cambria}]{ma2020survey}
Yukun Ma, Khanh~Linh Nguyen, Frank~Z Xing, and Erik Cambria. 2020.
\newblock A survey on empathetic dialogue systems.
\newblock \emph{Information Fusion}, 64:50--70.

\bibitem[{Mesgar et~al.(2020)Mesgar, Simpson, and Gurevych}]{mesgar2020improving}
Mohsen Mesgar, Edwin Simpson, and Iryna Gurevych. 2020.
\newblock Improving factual consistency between a response and persona facts.
\newblock \emph{arXiv preprint arXiv:2005.00036}.

\bibitem[{Myers(1962)}]{myers1962myers}
Isabel~Briggs Myers. 1962.
\newblock The myers-briggs type indicator: Manual (1962).

\bibitem[{OpenAI(2022)}]{chatgpt}
OpenAI. 2022.
\newblock \href {https://openai.com/blog/chatgpt} {Openai: Introducing chatgpt}.

\bibitem[{Pan and Zeng(2023)}]{pan2023llms}
Keyu Pan and Yawen Zeng. 2023.
\newblock Do llms possess a personality? making the mbti test an amazing evaluation for large language models.
\newblock \emph{arXiv preprint arXiv:2307.16180}.

\bibitem[{Park et~al.(2023)Park, O'Brien, Cai, Morris, Liang, and Bernstein}]{park2023generative}
Joon~Sung Park, Joseph O'Brien, Carrie~Jun Cai, Meredith~Ringel Morris, Percy Liang, and Michael~S Bernstein. 2023.
\newblock Generative agents: Interactive simulacra of human behavior.
\newblock In \emph{Proceedings of the 36th Annual ACM Symposium on User Interface Software and Technology}, pages 1--22.

\bibitem[{Pearson(1901)}]{pearson1901liii}
Karl Pearson. 1901.
\newblock Liii. on lines and planes of closest fit to systems of points in space.
\newblock \emph{The London, Edinburgh, and Dublin philosophical magazine and journal of science}, 2(11):559--572.

\bibitem[{Rawte et~al.(2023)Rawte, Sheth, and Das}]{rawte2023survey}
Vipula Rawte, Amit Sheth, and Amitava Das. 2023.
\newblock A survey of hallucination in large foundation models.
\newblock \emph{arXiv preprint arXiv:2309.05922}.

\bibitem[{Shao et~al.(2023)Shao, Li, Dai, and Qiu}]{shao2023character}
Yunfan Shao, Linyang Li, Junqi Dai, and Xipeng Qiu. 2023.
\newblock Character-llm: A trainable agent for role-playing.
\newblock \emph{arXiv preprint arXiv:2310.10158}.

\bibitem[{Silver et~al.(2022)Silver, Hariprasad, Shuttleworth, Kumar, Lozano-P{\'e}rez, and Kaelbling}]{silver2022pddl}
Tom Silver, Varun Hariprasad, Reece~S Shuttleworth, Nishanth Kumar, Tom{\'a}s Lozano-P{\'e}rez, and Leslie~Pack Kaelbling. 2022.
\newblock Pddl planning with pretrained large language models.
\newblock In \emph{NeurIPS 2022 Foundation Models for Decision Making Workshop}.

\bibitem[{Song et~al.(2023)Song, Wu, Washington, Sadler, Chao, and Su}]{song2023llm}
Chan~Hee Song, Jiaman Wu, Clayton Washington, Brian~M Sadler, Wei-Lun Chao, and Yu~Su. 2023.
\newblock Llm-planner: Few-shot grounded planning for embodied agents with large language models.
\newblock In \emph{Proceedings of the IEEE/CVF International Conference on Computer Vision}, pages 2998--3009.

\bibitem[{Tu et~al.(2023)Tu, Chen, Li, Li, Shang, Zhao, Wang, and Yan}]{tu2023characterchat}
Quan Tu, Chuanqi Chen, Jinpeng Li, Yanran Li, Shuo Shang, Dongyan Zhao, Ran Wang, and Rui Yan. 2023.
\newblock Characterchat: Learning towards conversational ai with personalized social support.
\newblock \emph{arXiv preprint arXiv:2308.10278}.

\bibitem[{Vaswani et~al.(2017)Vaswani, Shazeer, Parmar, Uszkoreit, Jones, Gomez, Kaiser, and Polosukhin}]{vaswani2017attention}
Ashish Vaswani, Noam Shazeer, Niki Parmar, Jakob Uszkoreit, Llion Jones, Aidan~N Gomez, {\L}ukasz Kaiser, and Illia Polosukhin. 2017.
\newblock Attention is all you need.
\newblock \emph{Advances in neural information processing systems}, 30.

\bibitem[{Wang et~al.(2023{\natexlab{a}})Wang, Ma, Feng, Zhang, Yang, Zhang, Chen, Tang, Chen, Lin et~al.}]{wang2023survey}
Lei Wang, Chen Ma, Xueyang Feng, Zeyu Zhang, Hao Yang, Jingsen Zhang, Zhiyuan Chen, Jiakai Tang, Xu~Chen, Yankai Lin, et~al. 2023{\natexlab{a}}.
\newblock A survey on large language model based autonomous agents.
\newblock \emph{arXiv preprint arXiv:2308.11432}.

\bibitem[{Wang et~al.(2023{\natexlab{b}})Wang, Fei, Leng, and Li}]{wang2023does}
Xintao Wang, Yaying Fei, Ziang Leng, and Cheng Li. 2023{\natexlab{b}}.
\newblock Does role-playing chatbots capture the character personalities? assessing personality traits for role-playing chatbots.
\newblock \emph{arXiv preprint arXiv:2310.17976}.

\bibitem[{Wang et~al.(2022)Wang, Wei, Schuurmans, Le, Chi, Narang, Chowdhery, and Zhou}]{wang2022self}
Xuezhi Wang, Jason Wei, Dale Schuurmans, Quoc Le, Ed~Chi, Sharan Narang, Aakanksha Chowdhery, and Denny Zhou. 2022.
\newblock Self-consistency improves chain of thought reasoning in language models.
\newblock \emph{arXiv preprint arXiv:2203.11171}.

\bibitem[{Wang et~al.(2023{\natexlab{c}})Wang, Peng, Que, Liu, Zhou, Wu, Guo, Gan, Ni, Zhang et~al.}]{wang2023rolellm}
Zekun~Moore Wang, Zhongyuan Peng, Haoran Que, Jiaheng Liu, Wangchunshu Zhou, Yuhan Wu, Hongcheng Guo, Ruitong Gan, Zehao Ni, Man Zhang, et~al. 2023{\natexlab{c}}.
\newblock Rolellm: Benchmarking, eliciting, and enhancing role-playing abilities of large language models.
\newblock \emph{arXiv preprint arXiv:2310.00746}.

\bibitem[{Wei et~al.(2022)Wei, Wang, Schuurmans, Bosma, Xia, Chi, Le, Zhou et~al.}]{wei2022chain}
Jason Wei, Xuezhi Wang, Dale Schuurmans, Maarten Bosma, Fei Xia, Ed~Chi, Quoc~V Le, Denny Zhou, et~al. 2022.
\newblock Chain-of-thought prompting elicits reasoning in large language models.
\newblock \emph{Advances in Neural Information Processing Systems}, 35:24824--24837.

\bibitem[{Xue et~al.(2023)Xue, Wang, Wang, Mi, Wang, Wang, Shang, Jiang, Liu, and Wong}]{xue2023improving}
Boyang Xue, Weichao Wang, Hongru Wang, Fei Mi, Rui Wang, Yasheng Wang, Lifeng Shang, Xin Jiang, Qun Liu, and Kam-Fai Wong. 2023.
\newblock Improving factual consistency for knowledge-grounded dialogue systems via knowledge enhancement and alignment.
\newblock \emph{arXiv preprint arXiv:2310.08372}.

\bibitem[{Yan et~al.(2022)Yan, Li, Yu et~al.}]{yan2022deep}
Rui Yan, Juntao Li, Zhou Yu, et~al. 2022.
\newblock Deep learning for dialogue systems: Chit-chat and beyond.
\newblock \emph{Foundations and Trends{\textregistered} in Information Retrieval}, 15(5):417--589.

\bibitem[{Zhang et~al.(2021)Zhang, Chen, D'Haro, Zhang, Friedrichs, Lee, and Li}]{zhang2021dynaeval}
Chen Zhang, Yiming Chen, Luis~Fernando D'Haro, Yan Zhang, Thomas Friedrichs, Grandee Lee, and Haizhou Li. 2021.
\newblock Dynaeval: Unifying turn and dialogue level evaluation.
\newblock \emph{arXiv preprint arXiv:2106.01112}.

\bibitem[{Zhang et~al.(2023)Zhang, Li, Cui, Cai, Liu, Fu, Huang, Zhao, Zhang, Chen et~al.}]{zhang2023siren}
Yue Zhang, Yafu Li, Leyang Cui, Deng Cai, Lemao Liu, Tingchen Fu, Xinting Huang, Enbo Zhao, Yu~Zhang, Yulong Chen, et~al. 2023.
\newblock Siren's song in the ai ocean: A survey on hallucination in large language models.
\newblock \emph{arXiv preprint arXiv:2309.01219}.

\bibitem[{Zhao et~al.(2023)Zhao, Zhou, Li, Tang, Wang, Hou, Min, Zhang, Zhang, Dong et~al.}]{zhao2023survey}
Wayne~Xin Zhao, Kun Zhou, Junyi Li, Tianyi Tang, Xiaolei Wang, Yupeng Hou, Yingqian Min, Beichen Zhang, Junjie Zhang, Zican Dong, et~al. 2023.
\newblock A survey of large language models.
\newblock \emph{arXiv preprint arXiv:2303.18223}.

\bibitem[{Zhao et~al.(2020)Zhao, Wu, Tao, Xu, Zhao, and Yan}]{zhao2020low}
Xueliang Zhao, Wei Wu, Chongyang Tao, Can Xu, Dongyan Zhao, and Rui Yan. 2020.
\newblock Low-resource knowledge-grounded dialogue generation.
\newblock \emph{arXiv preprint arXiv:2002.10348}.

\bibitem[{Zheng et~al.(2023)Zheng, Chiang, Sheng, Zhuang, Wu, Zhuang, Lin, Li, Li, Xing et~al.}]{zheng2023judging}
Lianmin Zheng, Wei-Lin Chiang, Ying Sheng, Siyuan Zhuang, Zhanghao Wu, Yonghao Zhuang, Zi~Lin, Zhuohan Li, Dacheng Li, Eric Xing, et~al. 2023.
\newblock Judging llm-as-a-judge with mt-bench and chatbot arena.
\newblock \emph{arXiv preprint arXiv:2306.05685}.

\bibitem[{Zheng et~al.(2019)Zheng, Chen, Huang, Liu, and Zhu}]{zheng2019personalized}
Yinhe Zheng, Guanyi Chen, Minlie Huang, Song Liu, and Xuan Zhu. 2019.
\newblock Personalized dialogue generation with diversified traits.
\newblock \emph{arXiv preprint arXiv:1901.09672}.

\bibitem[{Zheng et~al.(2020)Zheng, Zhang, Huang, and Mao}]{zheng2020pre}
Yinhe Zheng, Rongsheng Zhang, Minlie Huang, and Xiaoxi Mao. 2020.
\newblock A pre-training based personalized dialogue generation model with persona-sparse data.
\newblock In \emph{Proceedings of the AAAI Conference on Artificial Intelligence}, volume~34, pages 9693--9700.

\bibitem[{Zhong et~al.(2022)Zhong, Dou, Zhu, Qian, and Wen}]{zhong2022less}
Hanxun Zhong, Zhicheng Dou, Yutao Zhu, Hongjin Qian, and Ji-Rong Wen. 2022.
\newblock Less is more: Learning to refine dialogue history for personalized dialogue generation.
\newblock \emph{arXiv preprint arXiv:2204.08128}.

\bibitem[{Zhou et~al.(2023)Zhou, Chen, Wan, Wen, Song, Yu, Huang, Peng, Yang, Xiao et~al.}]{zhou2023characterglm}
Jinfeng Zhou, Zhuang Chen, Dazhen Wan, Bosi Wen, Yi~Song, Jifan Yu, Yongkang Huang, Libiao Peng, Jiaming Yang, Xiyao Xiao, et~al. 2023.
\newblock Characterglm: Customizing chinese conversational ai characters with large language models.
\newblock \emph{arXiv preprint arXiv:2311.16832}.

\end{thebibliography}
\bibliographystyle{acl_natbib}

\appendix
\section{Evaluation Result by GPT-4}
Although GPT-4 has demonstrated the self-enhancement bias~\cite{zheng2023judging} and has an lower correlation with human judgement~\ref{table:coefficients}, we present the evaluation result by GPT-4 in 2-shot setting for reference, as shown in Table~\ref{tab:gpt4results}.

\begin{table*}[!h]
\centering
\small

\begin{tabular}{|l|>{\centering\arraybackslash}m{0.91cm}>{\centering\arraybackslash}m{0.91cm}>{\centering\arraybackslash}m{0.91cm}>{\centering\arraybackslash}m{0.91cm}>{\centering\arraybackslash}m{0.91cm}>{\centering\arraybackslash}m{0.91cm}|>{\centering\arraybackslash}m{1.8cm}|}
\toprule
                   & \multicolumn{6}{c|}{\textbf{Character Consistency}}                                                 & \textbf{Personality} \\ \cline{2-7} 
                   & KE             & KA             & KH             & PB             & PU             & Avg.           & \textbf{Back-Testing}         \\ \hline
ChatGLM3-6B        & 4.437          & 4.411          & 4.175          & 4.462          & 4.431          & 4.383          & 0.532                \\
XVERSE-7B          & 4.498          & 4.655          & 4.533          & 4.593          & 4.651          & 4.586          & 0.62                 \\
Baichuan2-7B       & 4.506          & 4.665          & 4.531          & 4.633          & 4.686          & 4.604          & 0.625                \\
Qwen-7B            & 4.303          & 4.375          & 4.257          & 4.415          & 4.413          & 4.353          & 0.606                \\
InternLM-7B        & 4.367          & 4.497          & 4.403          & 4.454          & 4.638          & 4.472          & 0.63                 \\
XVERSE-13B         & \underline{4.709}    & 4.812          & 4.611          & 4.743          & 4.802          & 4.735          & 0.63                 \\
Baichuan2-13B      & 4.672          & \underline{4.841}    & \underline{4.733}    & 4.771          & 4.812          & \underline{4.766}    & 0.639                \\
Qwen-14B           & 4.637          & 4.644          & 4.530          & 4.674          & 4.688          & 4.635          & 0.62                 \\
InternLM-20B       & 4.699          & 4.734          & 4.568          & 4.676          & 4.735          & 4.682          & 0.648                \\
CharacterGLM       & 4.157          & 4.679          & 4.450          & 4.495          & 4.640          & 4.484          & -                    \\
Xingchen           & 4.366          & 4.638          & 4.488          & 4.650          & 4.704          & 4.569          & 0.63                 \\
MiniMax            & 4.692          & 4.827          & 4.674          & \underline{4.776}    & \underline{4.849}    & 4.763          & \underline{0.685}          \\
BC-NPC-Turbo & 4.478          & 4.811          & 4.655          & 4.730          & 4.833          & 4.701          & 0.681                \\
GPT-3.5            & 3.793          & 3.858          & 3.549          & 3.837          & 3.866          & 3.781          & 0.653                \\
GPT-4              & \textbf{4.924} & \textbf{4.923} & \textbf{4.899} & \textbf{4.912} & \textbf{4.906} & \textbf{4.913} & \textbf{0.694}      
\end{tabular}

\begin{tabular}{|l|cccccclcl|}
\midrule

                   & \multicolumn{4}{c|}{\textbf{Conversational Ability}}                                                                                      & \multicolumn{5}{c|}{\textbf{Role-playing Attractiveness}}                                                                                                                      \\ \cline{2-10} 
                   & Flu.                            & Coh.                            & Cons.                           & \multicolumn{1}{c|}{Avg.}           & HL                              & CS                              & \multicolumn{1}{c}{ED}             & Emp.                            & \multicolumn{1}{c|}{Avg.}           \\ \hline
ChatGLM3-6B        & \multicolumn{1}{l}{4.160}       & \multicolumn{1}{l}{4.552}       & \multicolumn{1}{l}{4.182}       & \multicolumn{1}{l|}{4.298}          & \multicolumn{1}{l}{4.360}       & \multicolumn{1}{l}{3.620}       & 3.410                              & \multicolumn{1}{l}{3.570}       & 3.740                               \\
XVERSE-7B          & \multicolumn{1}{l}{4.591}       & \multicolumn{1}{l}{4.725}       & \multicolumn{1}{l}{4.392}       & \multicolumn{1}{l|}{4.569}          & \multicolumn{1}{l}{4.601}       & \multicolumn{1}{l}{3.608}       & 3.331                              & \multicolumn{1}{l}{3.535}       & 3.769                               \\
Baichuan2-7B       & \multicolumn{1}{l}{4.636}       & \multicolumn{1}{l}{4.760}       & \multicolumn{1}{l}{4.596}       & \multicolumn{1}{l|}{4.664}          & \multicolumn{1}{l}{4.608}       & \multicolumn{1}{l}{3.497}       & 3.240                              & \multicolumn{1}{l}{3.610}       & 3.739                               \\
Qwen-7B            & \multicolumn{1}{l}{4.201}       & \multicolumn{1}{l}{4.540}       & \multicolumn{1}{l}{4.025}       & \multicolumn{1}{l|}{4.255}          & \multicolumn{1}{l}{4.333}       & \multicolumn{1}{l}{3.606}       & 3.362                              & \multicolumn{1}{l}{3.379}       & 3.670                               \\
InternLM-7B        & \multicolumn{1}{l}{4.468}       & \multicolumn{1}{l}{4.599}       & \multicolumn{1}{l}{4.189}       & \multicolumn{1}{l|}{4.418}          & \multicolumn{1}{l}{4.420}       & \multicolumn{1}{l}{3.396}       & 3.075                              & \multicolumn{1}{l}{3.312}       & 3.551                               \\
XVERSE-13B         & \multicolumn{1}{l}{4.708}       & \multicolumn{1}{l}{4.812}       & \multicolumn{1}{l}{4.559}       & \multicolumn{1}{l|}{4.693}          & \multicolumn{1}{l}{\underline{4.736}} & \multicolumn{1}{l}{\underline{3.736}} & \underline{3.533}                        & \multicolumn{1}{l}{\underline{3.758}} & \underline{3.941}                         \\
Baichuan2-13B      & \multicolumn{1}{l}{\underline{4.724}} & \multicolumn{1}{l}{\underline{4.847}} & \multicolumn{1}{l}{\underline{4.631}} & \multicolumn{1}{l|}{\textbf{4.734}} & \multicolumn{1}{l}{4.726}       & \multicolumn{1}{l}{3.559}       & 3.246                              & \multicolumn{1}{l}{3.670}       & 3.800                               \\
Qwen-14B           & \multicolumn{1}{l}{4.500}       & \multicolumn{1}{l}{4.758}       & \multicolumn{1}{l}{4.439}       & \multicolumn{1}{l|}{4.566}          & \multicolumn{1}{l}{4.613}       & \multicolumn{1}{l}{3.631}       & 3.531                              & \multicolumn{1}{l}{3.612}       & 3.847                               \\
InternLM-20B       & \multicolumn{1}{l}{4.497}       & \multicolumn{1}{l}{4.798}       & \multicolumn{1}{l}{4.579}       & \multicolumn{1}{l|}{4.625}          & \multicolumn{1}{l}{4.669}       & \multicolumn{1}{l}{3.559}       & 3.399                              & \multicolumn{1}{l}{3.602}       & 3.807                               \\
CharacterGLM       & 4.562                           & 4.538                           & 4.297                           & \multicolumn{1}{c|}{4.466}          & 4.429                           & 3.267                           & \multicolumn{1}{c}{2.931}          & 3.032                           & \multicolumn{1}{c|}{3.415}          \\
Xingchen           & 4.558                           & 4.677                           & 4.326                           & \multicolumn{1}{c|}{4.520}          & 4.584                           & 3.339                           & \multicolumn{1}{c}{3.076}          & 3.155                           & \multicolumn{1}{c|}{3.539}          \\
MiniMax            & \textbf{4.733}                  & 4.819                           & 4.580                           & \multicolumn{1}{c|}{4.710}          & 4.735                           & 3.511                           & \multicolumn{1}{c}{3.304}          & 3.557                           & \multicolumn{1}{c|}{3.777}          \\
BC-NPC-Turbo & 4.685                           & 4.770                           & 4.452                           & \multicolumn{1}{c|}{4.636}          & 4.581                           & 3.437                           & \multicolumn{1}{c}{3.157}          & 3.355                           & \multicolumn{1}{c|}{3.633}          \\
GPT-3.5            & 3.656                           & 3.788                           & 3.873                           & \multicolumn{1}{c|}{3.772}          & 3.710                           & 3.162                           & \multicolumn{1}{c}{2.795}          & 3.251                           & \multicolumn{1}{c|}{3.230}          \\
GPT-4              & 4.630                           & \textbf{4.850}                  & \textbf{4.656}                  & \multicolumn{1}{c|}{\underline{4.712}}    & \textbf{4.796}                  & \textbf{3.947}                  & \multicolumn{1}{c}{\textbf{3.806}} & \textbf{3.883}                  & \multicolumn{1}{c|}{\textbf{4.108}} \\ \bottomrule
\end{tabular}
\caption{Detailed evaluation results on \textit{CharacterEval}. The 12 subjective metrics in conversational ability, character consistency and role-playing attractiveness dimensions are evaluated by GPT-4. The best performances are highlighted in \textbf{bold}, while sub-optimal ones are marked with \underline{underline}. It is notable that the score for CharacterGLM in personality back-testing is unavailable, hence it is replaced by a ``-''.}
\label{tab:gpt4results}
\end{table*}


\end{document}